\documentclass[runningheads]{llncs}

\usepackage{eccv}

\usepackage{geometry}
\usepackage{eccvabbrv}  

\usepackage{graphicx}
\usepackage{pifont}
\usepackage{multirow}
\usepackage[table]{xcolor}
\usepackage{booktabs}
\usepackage{enumitem}
\newcommand{\cmark}{\ding{51}} \newcommand{\xmark}{\ding{55}}
\usepackage{pifont}
\usepackage[accsupp]{axessibility}  % Improves PDF readability for those with disabilities.
\usepackage{booktabs}
\usepackage{makecell}

\usepackage{hyperref}

\usepackage{orcidlink}

\begin{document}

\title{GridVAD: Open-Set Video Anomaly Detection via Spatial Reasoning over Stratified Frame Grids}
\titlerunning{GridVAD}
% Include the authors' ORCID for the camera-ready version, if at all possible.
\author{Mohamed Eltahir \inst{1} \orcidlink{0009-0003-5811-5978} \and
Ahmed O. Ibrahim $^*$ \inst{2} \orcidlink{0009-0005-8016-2978} \and
\mbox{Obada Siralkhatim $^*$} \inst{3} \orcidlink{0009-0006-8173-3506} \and \\
Tabarak Abdallah \inst{3} \orcidlink{0009-0002-7125-6313} \and
Sondos Mohamed $^\dagger$ \inst{4} \orcidlink{0009-0002-5647-6895} }

\authorrunning{M. Eltahir et al.}

\institute{King Abdullah University of Science and Technology (KAUST), Thuwal, Saudi Arabia \\
\email{mohamed.hamid@kaust.edu.sa}
\and
African Institute for Mathematical Sciences (AIMS), Cape Town, South Africa \\
\email{ahmed.o.a.ibrahim@aims.ac.za}
\and
Independent Researcher \\
\email{\{obadabadee.pro, tabarak.m.elgady\}@gmail.com}
\and
Faculty of Computers and Information Technology, University of Tabuk, Tabuk 71491, KSA \\
\email{s\_mohamed@ut.edu.sa}}


% First names are abbreviated in the running head

\maketitle
\renewcommand{\thefootnote}{}
\footnotetext{\scriptsize $^*$ Equal Contribution \quad $^\dagger$ Corresponding Author} 

\begin{abstract}

Vision-Language Models (VLMs) are powerful open-set reasoners, yet their direct use as
anomaly detectors in video surveillance is fragile: without calibrated anomaly priors,
they alternate between missed detections and hallucinated false alarms. We argue the
problem is not the VLM itself but how it is used. VLMs should function as
\emph{anomaly proposers}, generating open-set candidate descriptions that are then
grounded and tracked by purpose-built spatial and temporal modules.
We instantiate this \emph{propose-ground-propagate} principle in \textbf{\textit{GridVAD}}, a
training-free pipeline that produces pixel-level anomaly masks without any
domain-specific training. A VLM reasons over stratified grid representations of video
clips to generate natural-language anomaly proposals. Self-Consistency Consolidation
(SCC) filters hallucinations by retaining only proposals that recur across multiple
independent samplings. Grounding DINO anchors each surviving proposal to a bounding box,
and SAM2 propagates it as a dense mask through the anomaly interval. The per-clip model 
budget is fixed at $M{+}1$ calls regardless of video length, where $M$ can be set according to the proposals needed.
On UCSD Ped2, GridVAD achieves the highest Pixel-AUROC (77.59) among all compared
methods, surpassing even the partially fine-tuned TAO (75.11), and outperforms
other zero-shot approaches on object-level RBDC by over $5{\times}$. Ablations reveal
that SCC provides a controllable precision-recall tradeoff: filtering improves all
pixel-level metrics at a modest cost in object-level recall. Efficiency experiments
show GridVAD is $2.7{\times}$ more call-efficient than uniform per-frame VLM querying
while additionally producing dense segmentation masks. Code and qualitative video results are available at \url{https://gridvad.github.io}.

\end{abstract}

\section{Introduction}
\label{sec:intro}
% What is VAD and Why Is It Important?
Video Anomaly Detection (VAD) aims to detect and temporally localize events that deviate from established scene dynamics\cite{DBLP:conf/cvpr/HuangLZLL0L0G0D25}. Unlike closed-set action recognition, VAD is an inherently unbalanced, open-set problem, where anomalies are characterized by their rarity, diversity, and extreme context-dependency\cite{DBLP:conf/cvpr/ZanellaMM0024}. For instance, while running is considered a pedestrian regularity on a sidewalk, it may signify a critical security deviation within a restricted industrial facility. These properties make VAD particularly resistant to closed-vocabulary, distribution-fitting approaches, and have recently motivated the use of foundation models as a more flexible alternative. Despite these challenges, VAD remains indispensable in safety critical deployments, including autonomous driving \cite{DBLP:conf/cvpr/BogdollNZ22}, public transportation monitoring, and industrial surveillance\cite{DBLP:conf/cvpr/RothPZSBG22}, where reliable detection of anomalous events is essential to mitigate potential risks.

\medskip
Vision-Language Models (VLMs) appear ideally suited for VAD: they possess broad visual knowledge, can reason about novel event categories, and produce natural-language descriptions that are inherently interpretable. Early work (\textit{LAVAD}~\cite{DBLP:conf/cvpr/ZanellaMM0024}, \textit{SUVAD}~\cite{DBLP:conf/icassp/GaoYH25}) %\textit{Holmes-VAD}~\cite{DBLP:journals/corr/abs-2406-12235}) 
has demonstrated impressive zero-shot temporal detection using  LLM-derived frame scores. Yet these methods share %a common design choice: the VLM is used as a \emph{direct detector}, 
a common design choice: the VLM or LLM is directly employed to score individual frames or clips for anomaly probability. We argue that this direct-scoring paradigm is inherently fragile. VLMs and LLMs are trained on internet-scale data where anomalous events are neither rare nor systematically labeled, so their anomaly scores are not calibrated. The same model that correctly flags a vehicle in a pedestrian zone may also flag a person running, a shadow, or an empty scene, with comparable confidence. The result is a noisy signal that requires extensive post-processing to be actionable.

We propose that the correct role for a VLM in a VAD pipeline is not detection but \emph{proposal}: generating open-set natural-language descriptions of candidate anomalous events, without being required to make a calibrated binary decision. This separates two sub-problems that should not be conflated. The VLM handles what it does well: recognizing and describing unusual visual content in open vocabulary. Spatial grounding and temporal localization are then handled by purpose-built models (\textit{Grounding DINO~\cite{DBLP:conf/eccv/LiuZRLZYJLYSZZ24}, SAM2~\cite{DBLP:conf/iclr/RaviGHHR0KRRGMP25}}) that are designed for precise spatial reasoning. Reliability comes not from asking the VLM to be more accurate, but from filtering its outputs statistically: proposals that recur across multiple independent samplings of the same clip are retained, while one-off hallucinations are discarded.

We instantiate this \emph{propose-ground-propagate} principle in \textbf{\textit{GridVAD}},
a training free pipeline. A video clip is represented as a stratified spatial grid of frame samples, converting the temporal detection problem into a single-pass image understanding task. The VLM generates free-form anomaly proposals with temporal coordinates. A Self-Consistency Consolidation (SCC) stage semantically deduplicates proposals across $M$ independent grid samplings, retaining only those with cross-sampling support above a threshold. \textit{Grounding DINO} anchors each surviving proposal to a bounding box in the most informative frame, and \textit{SAM2} propagates it as a per-frame mask through the proposed interval. No component is trained on anomaly data. The pipeline is budget-controlled: regardless of video length, it requires exactly $M{+}1$ VLM calls per clip.

Experiments on \textit{UCSD Ped2}~\cite{DBLP:conf/cvpr/MahadevanLBV10} and \textit{ShanghaiTech Campus}~\cite{DBLP:conf/cvpr/LiuLLG18} validate this decomposition. On UCSD Ped2, \textbf{\textit{GridVAD}} achieves the highest Pixel-AUROC (77.59) among all compared methods, surpassing the partially fine-tuned TAO~\cite{DBLP:conf/cvpr/HuangLZLL0L0G0D25} (75.11), and outperforms other zero-shot approaches on object-level RBDC by over $5{\times}$, all without any dataset-specific fine-tuning or closed-set category lists. Ablations reveal a controllable precision-recall tradeoff governed by SCC, and efficiency experiments show GridVAD is $2.7{\times}$ more call-efficient than uniform per-frame VLM querying.

\medskip
\noindent\textbf{Contributions.}
\begin{itemize}[label=\textbullet]

\item We identify and articulate the \textbf{VLM-as-proposer design principle}: VLMs used as direct anomaly detectors produce noisy, uncalibrated outputs, but when repositioned as open-set proposers within a propose-ground-propagate decomposition, they enable reliable pixel-level anomaly localization without any task-specific training.

\item We introduce \textbf{Self-Consistency Consolidation (SCC)}, a lightweight hallucination filtering mechanism that treats the VLM as a stochastic sensor and retains only proposals with cross-sampling statistical support, requiring no learned parameters and no dataset-specific calibration. We show empirically that SCC controls a precision-recall tradeoff between pixel-level mask quality and object-level detection recall.

\item We propose the \textbf{stratified grid representation}, which converts temporal anomaly detection into a single-pass spatial reasoning task, enabling budget-controlled VLM querying that is independent of video length and $2.7{\times}$ more call-efficient than per-frame alternatives.
\end{itemize}

\section{Related Work}
\noindent\textbf{Training-based VAD.}
The dominant paradigm in VAD trains models to recognize normality and flag deviations, spanning fully supervised~\cite{DBLP:conf/mm/LiuM19,DBLP:conf/cvpr/WangYZHH19,DBLP:conf/cvpr/BaiHLWZSY19}, weakly supervised~\cite{DBLP:conf/cvpr/SultaniCS18,DBLP:conf/eccv/LiCZZ22,DBLP:journals/tip/WuL21,DBLP:conf/aaai/ZhouY023}, and one-class settings~\cite{DBLP:conf/cvpr/YangLWWL23,DBLP:conf/iccv/GongLLSMVH19,DBLP:conf/cvpr/SunG23,DBLP:conf/iccv/LiuNLZL21}. While effective within their target domains, these methods require annotated data or clean normality distributions, and they generalize poorly to anomaly categories not seen during training. Object-centric approaches~\cite{DBLP:conf/cvpr/IonescuKG019,DBLP:conf/cvpr/GeorgescuBIKPS21,DBLP:journals/tmlr/ReissH25} improve spatial precision by grounding anomaly scores in detected objects but remain bound to fixed detection vocabularies and per-dataset tuning.
\noindent\textbf{VLMs as direct anomaly detectors.}
Large Vision-Language Models (VLMs) have sparked interest in training-free VAD by enabling open-set semantic reasoning without distribution-fitting. \textit{LAVAD}~\cite{DBLP:conf/cvpr/ZanellaMM0024} computes frame-level anomaly scores by passing VLM-generated captions to an LLM conditioned on generic, persona-based context prompts. \textit{SUVAD}~\cite{DBLP:conf/icassp/GaoYH25} summarizes video content into natural language lists of normal and abnormal events, using an LLM to score scenes against these semantic definitions. However, because VLMs and LLMs are not trained on the skewed class distributions typical of surveillance videos, these zero-shot pipelines can be vulnerable to producing noisy and uncalibrated anomaly scores. To derive calibrated frame-level scores, \textit{Holmes-VAD}~\cite{DBLP:journals/corr/abs-2406-12235} and \textit{VERA}~\cite{DBLP:conf/cvpr/YeLH25} adapt VLMs through instruction fine-tuning and prompt optimization, respectively, but this comes at the cost of requiring domain-specific datasets or labeled video supervision. Furthermore, all these methods share a fundamental conceptual flaw: even after adaptation, the VLM or LLM is tasked with making a direct, end-to-end binary anomaly judgment at either the frame or clip level. This conflates two distinct sub-problems: (1) \emph{recognizing and describing} unusual visual content in open vocabulary, which VLMs excel at, and (2) \emph{deciding} whether that content constitutes an anomaly under long-tailed real-world distributions, which language-based scoring alone cannot handle robustly without heavy supervision or post-hoc heuristics. Indeed, all of the aforementioned VLM-based models are restricted to frame-level or clip-level prediction and do not provide fine-grained spatial grounding or object-level localization. A comprehensive comparison contrasting these existing VLM-based VAD methods with our proposed pipeline is provided in Table~\ref{tab:sota}.
\medskip

\noindent\textbf{Video Object Segmentation and Grounding.}
Segmenting and tracking target entities in dynamic video scenes is central to fine-grained video understanding. The Segment Anything Model (SAM)~\cite{DBLP:conf/iccv/KirillovMRMRGXW23} established promptable 2D mask generation, but was limited to static single-frame images. SAM 2~\cite{DBLP:conf/iclr/RaviGHHR0KRRGMP25} generalized this paradigm to video by integrating a streaming memory architecture to achieve temporal masklet propagation. Simultaneously, open-vocabulary promptable detectors such as YOLOE \cite{DBLP:journals/corr/abs-2503-07465}, YOLO-World~\cite{DBLP:journals/corr/abs-2401-17270}, and Grounding DINO~\cite{DBLP:conf/eccv/LiuZRLZYJLYSZZ24}  bridge natural language descriptions directly to spatial bounding boxes in real time. Recent foundation models like SAM 3~\cite{DBLP:journals/corr/abs-2511-16719} further integrate these paradigms for unified promptable concept segmentation. Despite their strong spatio-temporal tracking and real-time detection capabilities, these foundation models lack intrinsic awareness of event context or anomaly semantics in surveillance settings.
\medskip

\noindent\textbf{Object-centric grounding and tracking in VAD.} 
Integrating Video Object Segmentation (VOS) and grounding into VAD shifts the task from coarse frame-level classification to precise spatio-temporal localization. \textit{TAO}~\cite{DBLP:conf/cvpr/HuangLZLL0L0G0D25} moves beyond frame-level scoring by reframing VAD as object tracking: it combines a per-object anomaly scorer, a tracking-based box refinement module, and SAM 2~\cite{DBLP:conf/iclr/RaviGHHR0KRRGMP25} for pixel-level segmentation. This achieves strong localization but inherits fundamental constraints: the anomaly scorer relies on pose- and depth-derived features, requiring per-dataset fine-tuning, and the robustness filtering stage used by TAO depends on many hand-tuned hyperparameters. %The pipeline also operates frame-by-frame, incurring $\mathcal{O}(N)$ heavy inference calls per video.
\medskip

\noindent\textbf{Our position.}
\textbf{\textit{GridVAD}} uses a VLM to generate open-set natural-language proposals describing potential anomalous events without requiring binary anomaly decisions. A Self-Consistency Consolidation (SCC) filter retains proposals that recur across independent samplings, discarding one-off hallucinations. Purpose-built execution models then handle precise localization: \textit{Grounding DINO} provides spatial grounding for complex, long-tail text proposals, while \textit{SAM 2} leverages its streaming memory architecture for efficient, fine-grained spatio-temporal tracking. This modular decomposition requires no anomaly-domain training, no task-specific fine-tuning, and no hand-tuned threshold heuristics on the detection signal.

\begin{table}[t]
\centering
\small
\setlength{\tabcolsep}{0.9pt}
\caption{Comparison with recent VLM-based VAD methods. 
The adaptation strategy is denoted as \textit{FT} (fine-tuning), 
\textit{PT} (prompt-tuning), and \textit{TF} (training-free).}
\label{tab:sota}
%\begin{tabular}{lcccccc}
\begin{tabular}{lccccc}
\toprule
Method &
\makecell[c]{Open-set\\Reasoning} &
%\makecell[c]{Direct Binary\\Decision} &
\makecell[c]{Object-level\\Localization} &
\makecell[c]{Temporal\\Localization} &
\makecell[c]{Hallucination\\Filtering} &
\makecell[c]{Adaptation\\Strategy} \\
\midrule
LAVAD~\cite{DBLP:conf/cvpr/ZanellaMM0024}        & \cmark & \xmark & \xmark & \xmark & TF \\
SUVAD~\cite{DBLP:conf/icassp/GaoYH25}        & \cmark  & \xmark & \xmark & \cmark & TF \\
Holmes-VAD~\cite{DBLP:journals/corr/abs-2406-12235}   & \cmark & \xmark & \xmark & \xmark & FT \\
VERA~\cite{DBLP:conf/cvpr/YeLH25}         & \cmark & \xmark & \xmark & \xmark & PT \\
TAO~\cite{DBLP:conf/cvpr/HuangLZLL0L0G0D25}          & \xmark & \cmark  & \cmark & \xmark & FT \\
\midrule
\textbf{GridVAD (Ours)}
             & \cmark  & \cmark & \cmark & \cmark & TF \\
\bottomrule
\end{tabular}
\end{table}

%%%%%%%%%%%%%%%%%%%%%%%%%%%%%%%%%%%%%%%%%%%%%%%%%%%%%%%%%%%%%%%%%%%%%%%%5555

\subsection{Problem Formulation}
\label{sec:problem}

Let $\mathcal{V} = \{f_1, f_2, \ldots, f_N\}$ be a surveillance video of $N$ frames,
where each frame $f_t \in \mathbb{R}^{H \times W \times 3}$ at time index $t \in \{1, \dots, N\}$ has Height $H$, Width $W$, and 3 color Channels.
We define a \emph{video anomaly} as an unexpected event that (i) occupies a
contiguous temporal interval $[t_s, t_e] \subset [1,N]$, where $t_s$ and $t_e$ denote the start and end frame indices, respectively; and
(ii) manifests as a spatially localized region in the affected frames.
The goal of fine-grained VAD is to jointly produce, \emph{without any
predefined category list or domain-specific training}:
\begin{enumerate}
  \item a natural-language description $d_j$ of each detected anomaly $j$,
  \item its temporal extent $[t_s, t_e]$,
  \item per-frame binary segmentation masks $\{S_t\}_{t=t_s}^{t_e}$, where $S_t \in \{0,1\}^{H\times W}$.
 % \item  per-frame binary segmentation masks $\{S_t^j\}_{t=t_s}^{t_e}$, where $S_t^j \in \{0, 1\}^{H \times W}$ isolates anomaly instance $j$.
\end{enumerate}

% ------------------------------------------------------------------
\subsection{Stratified Grid Sampling}
\label{sec:grid}

A video clip $C = \{f_s, f_{s+1}, \dots, f_e\} \subseteq \mathcal{V}$ of length
$L = e - s + 1$ frames is processed as follows.
We fix a grid of size $K = g^2$ cells (we use $g=3$, so $K=9$).
The clip $C$ is divided into $K$ equal-width temporal bins:
\begin{equation}
  B_k = \Bigl[s + \tfrac{(k-1)L}{K},\; s + \tfrac{kL}{K}\Bigr],
  \quad k = 1, \ldots, K.
  \label{eq:bins}
\end{equation}
To observe the clip across $M$ independent grid samplings, for each grid index $m \in \{1, \ldots, M\}$, one frame index
$i_k^{(m)} \sim \mathrm{Uniform}(B_k)$ is drawn independently from
each bin $B_k$, and the $K$ retrieved frames are tiled into a single
composite grid image:
\begin{equation}
  \mathcal{G}^{(m)} = \mathrm{Tile}\!\bigl(f_{i_1^{(m)}},\ldots,f_{i_K^{(m)}}\bigr)
  \in \mathbb{R}^{gH \times gW \times 3}.
  \label{eq:grid}
\end{equation}
Stratified sampling guarantees every temporal bin is represented in each
grid while the randomness ensures that the $M$ grids show
\emph{different frames} from each bin, providing complementary views
of the same clip at negligible computational cost.
Our approach
examines only $M {\times} K$ frames in total, with each of the
$M$ VLM calls processing an entire clip's worth of temporal context
in a \emph{single forward pass}.

% ------------------------------------------------------------------

\begin{figure*}[t]
\centering
\includegraphics[width=0.88\textwidth, trim=2mm 2mm 1mm 1mm, clip]{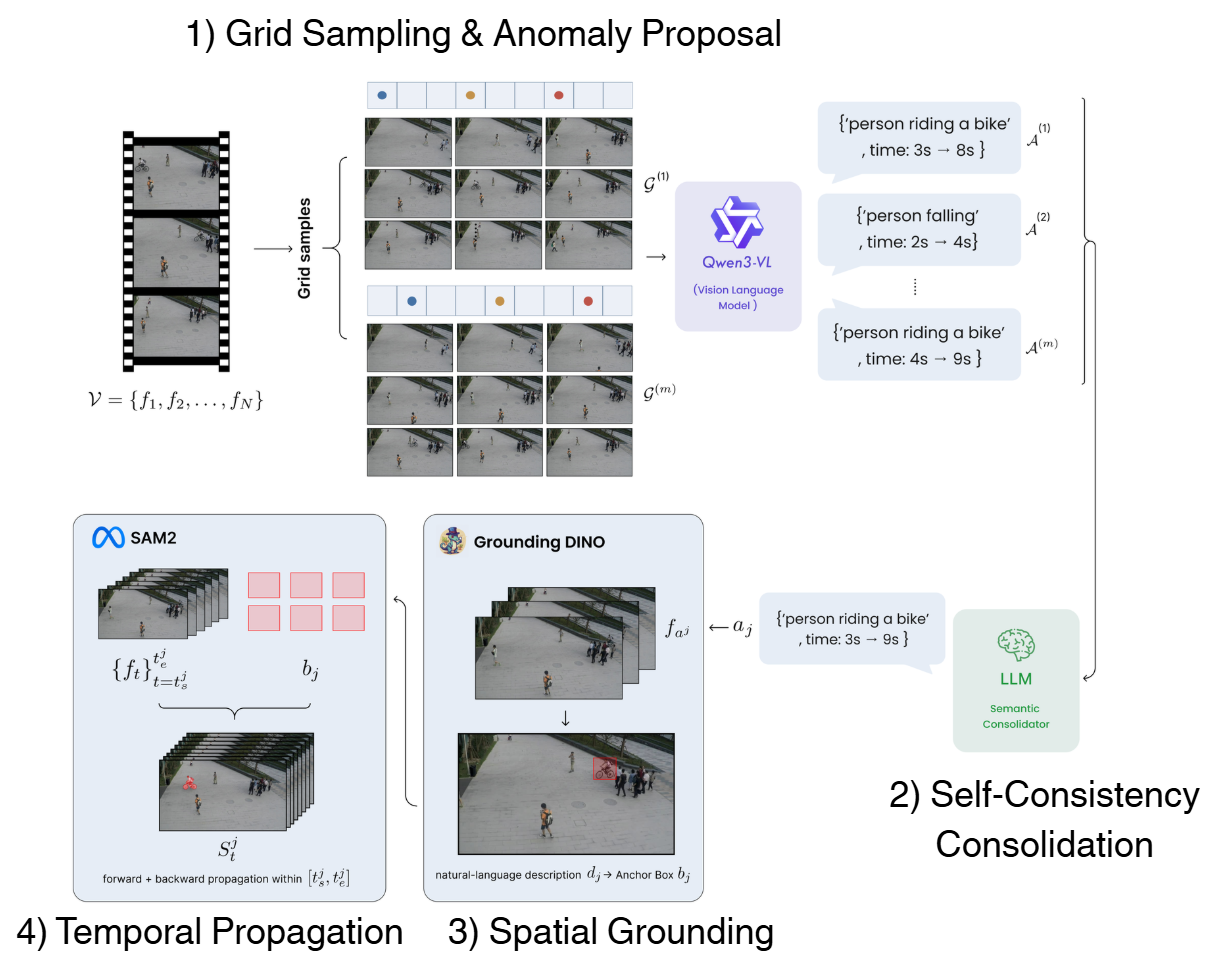}
\caption{\textbf {Overview of the \textit{GridVAD}} pipeline. \textit{GridVAD} decomposes training-free video anomaly detection into a four-stage pipeline: 
\textbf{(1) Grid Sampling \& Anomaly Proposal:} A video clip $\mathcal{V}$ is converted into $M$ stratified frame grids $\{\mathcal{G}^{(m)}\}_{m=1}^M$ and processed the VLM to generate $M$ open-set anomaly proposal sets $\{\mathcal{A}^{(m)}\}_{m=1}^M$. 
\textbf{(2) Self-Consistency Consolidation:} An LLM filters hallucinatory proposals across samples to retain statistically consistent anomaly candidates $a_j$. 
\textbf{(3) Spatial Grounding:} \textit{Grounding DINO} anchors candidate description $d_j$ to a spatial bounding box $b_j$ in keyframe $f_{a_j}$. 
\textbf{(4) Temporal Propagation:} \textit{SAM2} propagates the mask $S_t^j$ bidirectionally within interval $[t_s^j, t_e^j]$ to produce temporally consistent pixel-level anomaly maps.}
\label{fig:gridvad}
\end{figure*}

Each grid $\mathcal{G}^{(m)}$ is passed to a VLM $\Phi$ with a structured prompt
$\mathcal{P}$ requesting free-form detection, description, and temporal
localization:
\begin{equation}
  \mathcal{A}^{(m)} = \Phi\!\bigl(\mathcal{G}^{(m)},\,\mathcal{P}\bigr)
  = \bigl\{(d_j,\, t_s^j,\, t_e^j,\, p_j)\bigr\}_{j},
  \label{eq:vlm}
\end{equation}
where $\mathcal{A}^{(m)}$ is the set of raw proposals generated for grid $m$. Each proposal $a_j = (d_j, t_j^s, t_j^e, p_j)$ consists of:
\begin{itemize}
    \item $d_j$: a free-form natural language description of the observed anomaly,
    \item $[t_j^s, t_j^e]$: the estimated temporal interval decoded from the VLM's reported grid cell indices via the bin-to-frame mapping in Eq.~(\ref{eq:bins}), and
    \item $p_j \in [0, 1]$: the VLM's self-reported confidence score.
\end{itemize}
The VLM is given \emph{no predefined anomaly category list}. It decides what constitutes an anomaly from first principles, enabling open-set detection of arbitrary events.

\subsection{Self-Consistency Consolidation (SCC)}
\label{sec:scc}
The $M$ sets of proposals $\{\mathcal{A}^{(m)}\}_{m=1}^M$ generated by the VLM across different grid samplings are pooled
into a candidate set and processed by a text-only LLM call $\Phi_{\text{text}}$ acting as a semantic consolidator:
\begin{equation}
  \mathcal{A}^{\star} = \Phi_{\text{text}}\!\Bigl(
    \textstyle\bigcup_{m=1}^{M} \mathcal{A}^{(m)},\;
    \mathcal{P}_{\text{SCC}}
  \Bigr).
  \label{eq:scc}
\end{equation}

Each consolidated entry $a_j \in \mathcal{A}^\star$ carries a \emph{support count} $\sigma_j \in \{1, \ldots, M\}$ recording how many of the $M$ samplings contributed a semantically matching proposal. To eliminate stochastic hallucinations, only entries satisfying a minimum consensus threshold $\tau$ are retained:
\begin{equation}
  \mathcal{A}^{\dagger} = \{ a_j \in \mathcal{A}^{\star} \mid \sigma_j \geq \tau \},
  \label{eq:filter}
\end{equation}
where $\mathcal{A}^\dagger$ denotes the final set of verified anomaly proposals. This consolidation step acts as the primary mechanism for converting noisy proposals into reliable detection signals: genuine anomalies recur across samplings ($\sigma_j \ge \tau$), whereas stochastic hallucinations are filtered out ($\sigma_j < \tau$).

The consolidation prompt $\mathcal{P}_{\text{SCC}}$ instructs the model to (i) group proposals
that refer to the \emph{same object instance} across different
samplings into a single entry $a_j$, (ii) merge their temporal intervals
by taking $t_s^\star = \min_m t_s^{(m)}$ and
$t_e^\star = \max_m t_e^{(m)}$, and (iii) keep genuinely distinct
anomalies as separate entries.
This is qualitatively different from heuristic IoU-based box filtering:
the VLM reasons about \emph{semantic identity}, e.g., a vehicle moving from
left-to-center across two grids is correctly recognized as one
continuous event, while a vehicle and a pedestrian in the same frames
remain separate.
The result is a de-duplication mechanism that requires
no score thresholds and no dataset-specific tuning.

\medskip
\noindent\textbf{Temporal precision and the role of $M$.}
Because each of the $M$ grids samples \emph{different} frames from the
same temporal bins, independent samplings observe the anomaly at
different moments within each bin of width $L/K$.
SCC merges them by union: the consolidated interval
$[t_s^\star, t_e^\star]$ therefore converges toward the true event
boundaries as $M$ grows, since each additional sampling provides an
independently-drawn observation from the boundary bins.

% ------------------------------------------------------------------
\subsection{Spatial Grounding and Pixel-Level Propagation}
\label{sec:sam2}

\noindent\textbf{Zero-shot spatial grounding.}
For each surviving proposal $a_j \in \mathcal{A}^{\dagger}$, the natural-language
description $d_j$ is passed to
Grounding DINO~\cite{DBLP:conf/eccv/LiuZRLZYJLYSZZ24}, a zero-shot open-vocabulary detector that localizes the described entity in a candidate anchor frame $f_{a_j}$:
\begin{equation}
    b_j = \mathrm{GroundingDINO}(f_{a_j},\; d_j),
    \label{eq:grounding}
\end{equation}
where $b_j$ is the predicted 2D bounding box prompt. The anchor frame $f_{a_j}$ is selected as the frame with the highest Grounding DINO confidence across $R$ candidates sampled from $[t_s^j, t_e^j]$. This step bridges the gap between the VLM's natural-language output and the pixel-level input required by SAM2, without any category-specific training.
\medskip

\noindent\textbf{Temporal mask propagation.} Given the anchor box $b_j$ in frame $f_{a_j}$, we initialize a \textit{SAM2} \cite{DBLP:conf/iclr/RaviGHHR0KRRGMP25} video propagation session with $b_j$ as the spatial prompt
and propagate both forward and backward within the temporal window
$[t_s^j, t_e^j]$:
\begin{equation}
  \{S_t^j\}_{t=t_s^j}^{t_e^j}
  \;=\;
  \mathrm{SAM2}\!\bigl(\{f_t\}_{t=t_s^j}^{t_e^j},\; b_j,\; a_j\bigr),
  \label{eq:sam2}
\end{equation}
where $S_t^j\in\{0,1\}^{H\times W}$ is the per-frame binary mask for proposal $j$.
The final anomaly mask for frame $t$ is the union over all detected
instances: $S_t = \bigcup_j S_t^j$.
Propagation is restricted to the SCC-refined window $[t_s^j, t_e^j]$,
so tracking errors cannot accumulate across unrelated video segments.

% ------------------------------------------------------------------
\subsection{Full Pipeline Summary and Complexity}
\label{sec:summary}
The complete pipeline is illustrated in Fig.~\ref{fig:gridvad} and summarized as follows:
\begin{align}
  \{\mathcal{G}^{(m)}\}_{m=1}^M &= \mathrm{Tile}(\mathcal{V};\, M)
     & &\text{[$\mathcal{O}(1)$ copy]} \label{eq:s1}\\
  \{\mathcal{A}^{(m)}\}_{m=1}^M &= \Phi(\{\mathcal{G}^{(m)}\}_{m=1}^M,\,\mathcal{P})
     & &\text{[$M$ VLM forward passes]} \label{eq:s2}\\
  \mathcal{A}^{\star} &= \mathrm{SCC}(\{\mathcal{A}^{(m)}\}_{m=1}^M)
     & &\text{[1 SCC pass]} \label{eq:scc_summary}\\
\mathcal{A}^{\dagger} &= \{ a_j \in \mathcal{A}^{\star} \mid \sigma_j \geq \tau \}
& &\text{[support filter]} \label{eq:filter_summary}\\
  b_j &= \mathrm{GroundingDINO}(f_{a_j},\, d_j)
     & &\text{[$|\mathcal{A}^{\dagger}|$ grounding calls]} \label{eq:s3}\\
  S_t^j &= \mathrm{SAM2}(\{f_t\}, b_j, a_j)
     & &\text{[$|\mathcal{A}^{\dagger}|$ propagation passes]} \label{eq:s4}
\end{align}
where $|\mathcal{A}^{\dagger}|$ is the number of detected anomalies. The dominant cost is the $M$ VLM calls~\eqref{eq:s2};
all subsequent steps process only the anomalous window.

\section{Experiments}
\subsection{Experimental Setup}

\noindent\textbf{Datasets.}
We evaluate on two standard VAD benchmarks. In both datasets, the training set consists exclusively of normal video events, whereas anomalies appear only within the test set.
\textit{ShanghaiTech Campus}~\cite{DBLP:conf/cvpr/LiuLLG18} contains 330 normal training videos and 107 test videos ($480{\times}856$ pixels) across 13 campus scenes, with test-set anomalies including fighting, robbery, and cycling in restricted areas.
\textit{UCSD Ped2}~\cite{DBLP:conf/cvpr/MahadevanLBV10} contains 16 normal training videos and 12 test videos ($240{\times}360$ pixels), with test-set anomalies including cyclists, vehicles, and skateboarders operating in pedestrian zones.

\noindent\textbf{Evaluation Metrics.}
We adopt the same evaluation protocol as \textit{TAO}~\cite{DBLP:conf/cvpr/HuangLZLL0L0G0D25}, spanning
three complementary granularities.

\emph{Frame-level.}
\textit{Frame-AUROC} measures the model's ability to distinguish anomalous from
normal frames, computed as the area under the
receiver operating characteristic curve using per-frame anomaly scores.
A score of 1.0 indicates perfect temporal discrimination.

\emph{Pixel-level.} %Following~\cite{DBLP:conf/cvpr/HuangLZLL0L0G0D25},
We evaluate four pixel-level metrics:
\textit{Pixel-AUROC} assesses the model’s capability to differentiate between normal and anomalous pixels across thresholds.
\textit{Pixel-AUPRO} evaluates segmentation accuracy based on region overlap.
\textit{Pixel-AP} evaluates detection accuracy by balancing false positives and false negatives.
%\textit{Pixel-AUPRO} reports the area under the precision--recall curve
%(PR-AUC) at the pixel level, following the same convention as TAO.
\textit{Pixel-F1} combines precision and recall into a single measure. 
All pixel-level metrics are computed per annotated frame and averaged over frames, then over videos.

\emph{Object-level.}
We report the Region-Based Detection Criterion (\textit{RBDC}) and
Track-Based Detection Criterion (\textit{TBDC}) introduced
in~\cite{DBLP:journals/pami/GeorgescuBIKPS22}, following the same
protocol as~\cite{DBLP:conf/cvpr/HuangLZLL0L0G0D25}.
\textit{RBDC} evaluates spatial accuracy based on intersection over union (IoU) with a threshold of $\alpha$.
\textit{TBDC} assesses temporal consistency in tracking anomalous regions over frames.
Both metrics are computed globally across the entire test set, as in \textit{TAO}~\cite{DBLP:conf/cvpr/HuangLZLL0L0G0D25}, rather than being averaged per video.

\emph{Anomaly score construction.}
For each verified proposal $a_j \in \mathcal{A}^{\dagger}$, the propagated mask
$S_t^j$ is weighted by the proposal's confidence $p_j$, and the pixel-level
anomaly map for frame $t$ is the pixel-wise maximum over proposals active at $t$:
\begin{equation}
  A_t(x,y) = \max_{j\,:\,t \in [t_s^j,\, t_e^j]} p_j \cdot S_t^j(x,y),
  \label{eq:pixelscore}
\end{equation}
with $A_t = 0$ where no proposal is active. Frame-level scores are the spatial
maximum of this map, $s_t = \max_{x,y} A_t(x,y)$. Because $S_t^j$ is binary,
$A_t$ takes a small number of distinct values, so threshold-swept metrics are
evaluated over a correspondingly coarse operating range.

\noindent\textbf{Implementation Details.}
%All experiments run on 4$\times$\textit{NVIDIA V100 32\,GB GPUs}.
%We used \emph{Qwen3-VL-30B-A3B-Instruct}~\cite{DBLP:journals/corr/abs-2511-21631} for both
%per-grid proposals and SCC consolidation, with support threshold $\tau{=}3$.
%Spatial grounding uses \emph{Grounding DINO}\cite{DBLP:conf/eccv/LiuZRLZYJLYSZZ24}
%with box and text thresholds $\delta{=}0.05$.
%Pixel-level tracking uses \emph{SAM2.1-Hiera-Tiny}\cite{DBLP:conf/iclr/RaviGHHR0KRRGMP25}
%with a minimum mask area of 50\,px$^2$.
%Each clip uses grid size $g{=}3$ ($K{=}9$ frames/grid) and $M{=}5$ independent
%stratified samplings. No component is fine-tuned on target datasets.
All experiments are conducted on  4$\times$\textit{NVIDIA V100 32\,GB GPUs}.
We use \emph{Qwen3-VL-30B-A3B-Instruct}~\cite{DBLP:journals/corr/abs-2511-21631} for both
per-grid proposal generation and SCC consolidation, setting the support threshold to $\tau = 3$.
Spatial grounding is performed using \emph{Grounding DINO}~\cite{DBLP:conf/eccv/LiuZRLZYJLYSZZ24}
with box and text confidence thresholds set to $\delta = 0.05$.
Pixel-level tracking uses \emph{SAM2.1-Hiera-Tiny}~\cite{DBLP:conf/iclr/RaviGHHR0KRRGMP25}
with a minimum mask area threshold of $50\,\text{px}^2$.
For each video clip, we set the grid dimension to $g = 3$ ($K = 9$ frames per grid) and sample $M = 5$ independent
stratified grids. No components are fine-tuned on the target datasets.
\subsection{Comparison with State-of-the-Art}
\label{sec:sota}

We compare \textit{GridVAD} against representative training-based and training-free baselines on
\textit{UCSD Ped2}~\cite{DBLP:conf/cvpr/MahadevanLBV10} and \textit{ShanghaiTech
Campus}~\cite{DBLP:conf/cvpr/LiuLLG18}.
As shown in Tables~\ref{tab:ped2_sota}--\ref{tab:shanghaitech_sota}, \textit{GridVAD} achieves the highest \textit{Pixel-AUROC} (77.59\%) on \textit{UCSD Ped2} among all compared methods, including the partially fine-tuned \textit{TAO} without any task-specific fine-tuning.
Furthermore, it delivers competitive performance across \textit{Pixel-AP}, \textit{Pixel-AUPRO}, and \textit{Pixel-F1}, though its \textit{RBDC} and \textit{TBDC} scores remain relatively low.

The gap in \textit{RBDC} and \textit{TBDC} reflects the proposal-recall bottleneck discussed in Section~\ref{sec:limitations}: \textit{GridVAD} can only ground anomalies that the VLM first proposes in the temporal montage, while exhaustive per-frame methods like \textit{TAO} track every object regardless. On \textit{ShanghaiTech Campus}, this recall gap is more pronounced due to diverse scenes and subtle anomalies.

\begin{table*}[t]
\centering
\caption{Quantitative comparison on \textit{UCSD Ped2}. Baseline numbers are transcribed from the
\textit{TAO} paper\cite{DBLP:conf/cvpr/HuangLZLL0L0G0D25}; our results are in the last row.}
\label{tab:ped2_sota}
\resizebox{\textwidth}{!}{
\begin{tabular}{lcccccc}
\toprule
\textbf{Method} & \textbf{Pixel-AUROC} & \textbf{Pixel-AP} & \textbf{Pixel-AUPRO} & \textbf{Pixel-F1} & \textbf{RBDC} & \textbf{TBDC} \\
\midrule
AdaCLIP (Fully fine-tuned)\cite{DBLP:conf/eccv/CaoZFCSB24} & 53.06 & 4.97 & 50.66 & 11.19 & 12.3 & 15.5 \\
AnomalyCLIP (Fully fine-tuned)\cite{DBLP:journals/corr/abs-2310-18961} & 54.25 & 23.73 & 38.59 & 7.48 & 13.1 & 21.0 \\
DDAD (Fully trained)\cite{DBLP:journals/corr/abs-2305-15956} & 55.87 & 5.61 & 15.12 & 2.67 & 18.01 & 13.29 \\
SimpleNet (Fully trained)\cite{DBLP:journals/corr/abs-2303-15140} & 52.49 & 20.51 & 44.05 & 10.71 & 51.18 & 27.75 \\
DRAEM (Fully trained)\cite{DBLP:conf/iccv/ZavrtanikKS21} & 69.58 & 30.63 & 35.78 & 10.89 & 44.26 & 70.64 \\
TAO (Partially fine-tuned)\cite{DBLP:conf/cvpr/HuangLZLL0L0G0D25} & 75.11 & \underline{50.78} & \underline{72.97} &\underline{64.12} &\underline{83.6} &\underline{ 93.2} \\
\rowcolor{gray!20}
AdaCLIP (Zero-shot)\cite{DBLP:conf/eccv/CaoZFCSB24}& 51.02 & 1.32 & 33.98 & 2.61 & 5.8 & 10.6 \\
\rowcolor{gray!20}
AnomalyCLIP (Zero-shot)\cite{DBLP:journals/corr/abs-2310-18961} & 51.63 & 21.20 & 36.34 & 5.92 & 7.5 & 11.2 \\
\midrule
\rowcolor{gray!20}
\textbf{GridVAD (Ours, Zero-shot)} & \underline{\textbf{77.59}} & \textbf{38.53} & \textbf{66.82} & \textbf{42.09} & \textbf{38.96} & \textbf{37.70} \\
\bottomrule
\end{tabular}
}
\end{table*}

\begin{table}[t]
\centering
\small
\setlength{\tabcolsep}{3pt}
\caption{Object-level comparison on \textit{ShanghaiTech Campus}. Baseline numbers are transcribed
from the \textit{TAO }paper\cite{DBLP:conf/cvpr/HuangLZLL0L0G0D25}.}
\label{tab:shanghaitech_sota}
\begin{tabular}{lcc}
\toprule
\textbf{Method} & \textbf{RBDC} & \textbf{TBDC} \\
\midrule
OCAD\cite{DBLP:conf/cvpr/IonescuKG019}& 20.7 & 44.5 \\
%BAF-AT & 41.3 & 78.8 \\
AED-SSMTL\cite{DBLP:conf/cvpr/GeorgescuBIKPS21} & 43.2 & 84.1 \\
HF2VAD\cite{DBLP:conf/iccv/LiuNLZL21} & 45.4 & 84.5 \\
STPT\cite{DBLP:conf/avss/NajiSLGA22} & 51.6 & 84.6 \\
TAO\cite{DBLP:conf/cvpr/HuangLZLL0L0G0D25} & \underline{62.1} & \underline{85.4} \\
\midrule
\rowcolor{gray!20}
\textbf{GridVAD (Ours)} &\textbf{33.32} & \textbf{14.58} \\
\bottomrule
\end{tabular}
\end{table}

\subsection{Ablation Study}
\label{sec:ablation}

\subsubsection{Does SCC Reduce VLM Hallucinations?}
\label{sec:abl_scc}

A core premise of \textit{GridVAD} is that VLMs are noisy proposers whose raw outputs
contain both genuine anomaly descriptions and confident-sounding hallucinations. Self-Consistency Consolidation (SCC) treats the VLM as a stochastic sensor and uses multiple independent samplings to separate signal from noise.
We isolate the effect of SCC by comparing $M{=}1$ (raw VLM output, no consolidation) against $M{=}5$ on a 15-video subset of ShanghaiTech Campus~\cite{DBLP:conf/cvpr/LiuLLG18}, keeping all other pipeline
components (\textit{Grounding DINO, SAM2}) fixed.

\begin{table}[t]
\centering
\small
\setlength{\tabcolsep}{4pt}
\caption{Effect of SCC on ShanghaiTech Campus (15 videos).
SCC improves all pixel-level metrics while reducing object-level recall,
revealing a precision-recall tradeoff.}
\label{tab:scc_ablation}
\begin{tabular}{lccccc}
\toprule
\textbf{Config} & \textbf{Px-AUROC} & \textbf{Px-AP} & \textbf{Px-F1} & \textbf{RBDC} & \textbf{TBDC} \\
\midrule
$M{=}1$, no SCC          & 62.71 & 25.36 & 29.64 & \textbf{32.31} & \textbf{33.18} \\
$M{=}5$ & \textbf{70.04} & \textbf{37.90} & \textbf{43.43} & 27.97 & 25.91 \\
\bottomrule
\end{tabular}
\end{table}

\noindent\textbf{Analysis.}
As shown in Table~\ref{tab:scc_ablation}, SCC produces a clear precision-recall tradeoff. The single-pass baseline ($M{=}1$) achieves higher \textit{RBDC} (+4.3) and \textit{TBDC} (+7.3) because it retains every VLM proposal, including hallucinations, some of which happen to overlap ground-truth regions. However, these unchecked proposals also produce noisy masks: \textit{Pixel-AUROC} drops by 7.3 points and \textit{Pixel-F1} by 13.8 points compared to the SCC-filtered configuration.

With $M{=}5$, SCC discards proposals that do not recur across independent samplings. This removes hallucinated detections and improves mask quality across all pixel-level metrics. The cost is a modest reduction in object-level recall, as some genuine but inconsistently proposed anomalies are also filtered. This tradeoff directly explains the gap between pixel-level and object-level scores observed in Section~\ref{sec:sota}:\textit{GridVAD} deliberately prioritizes detection precision over exhaustive recall. 

\subsection{Efficiency Paradigm Comparison}\label{sec:efficiency}
We compare two VLM querying strategies on UCSD Ped2. \textit{Uniform sampling}
queries the VLM once per sampled frame (every 10 frames), providing local detail
but no temporal context, with cost scaling linearly in video length.
\textit{GridVAD} (ours) batches $K{=}9$ frames into a spatial montage per call
and issues $M{=}5$ passes per clip, encoding both spatial and temporal context
with cost independent of video length.
As shown in Table~\ref{tab:efficiency}, uniform sampling achieves a higher
absolute Frame-AUROC (0.69 vs.\ 0.40) by dedicating each call to a single frame,
but requires $4.7{\times}$ more VLM calls and $2.3{\times}$ more wall time.
When normalized by VLM calls, GridVAD yields a higher Frame-AUROC per call
(0.0094 vs.\ 0.0035), suggesting that structured spatial batching can recover
useful detection signal per model invocation, though at reduced absolute
accuracy relative to dense per-frame coverage.

\begin{table}[t]
\centering
\small
\setlength{\tabcolsep}{5pt}
\caption{Efficiency comparison on \textit{UCSD Ped2}. \textit{GridVAD} uses
$4.7{\times}$ fewer VLM calls and runs $2.3{\times}$ faster, and additionally
produces dense pixel-level segmentation masks, though at lower absolute
Frame-AUROC.}
\label{tab:efficiency}
\begin{tabular}{lcccc}
\toprule
\textbf{Paradigm} & \textbf{VLM calls} & \textbf{Frame-AUROC} & \textbf{Time (s)} & \textbf{Fr-AUC / call} \\
\midrule
Uniform sampling        & 201 & 0.6948 & 640.6 & 0.0035 \\
\textbf{GridVAD (Ours)} & \textbf{43}  & 0.4024 & \textbf{276.7} & \textbf{0.0094} \\
\bottomrule
\end{tabular}
\end{table}
\subsection{Qualitative Analysis and Visualization}
\begin{figure*}[t]
\centering
\setlength{\tabcolsep}{6pt}
\renewcommand{\arraystretch}{1.0}
\makebox[\textwidth][c]{%
\begin{tabular}{c c}
\includegraphics[width=0.88\textwidth]{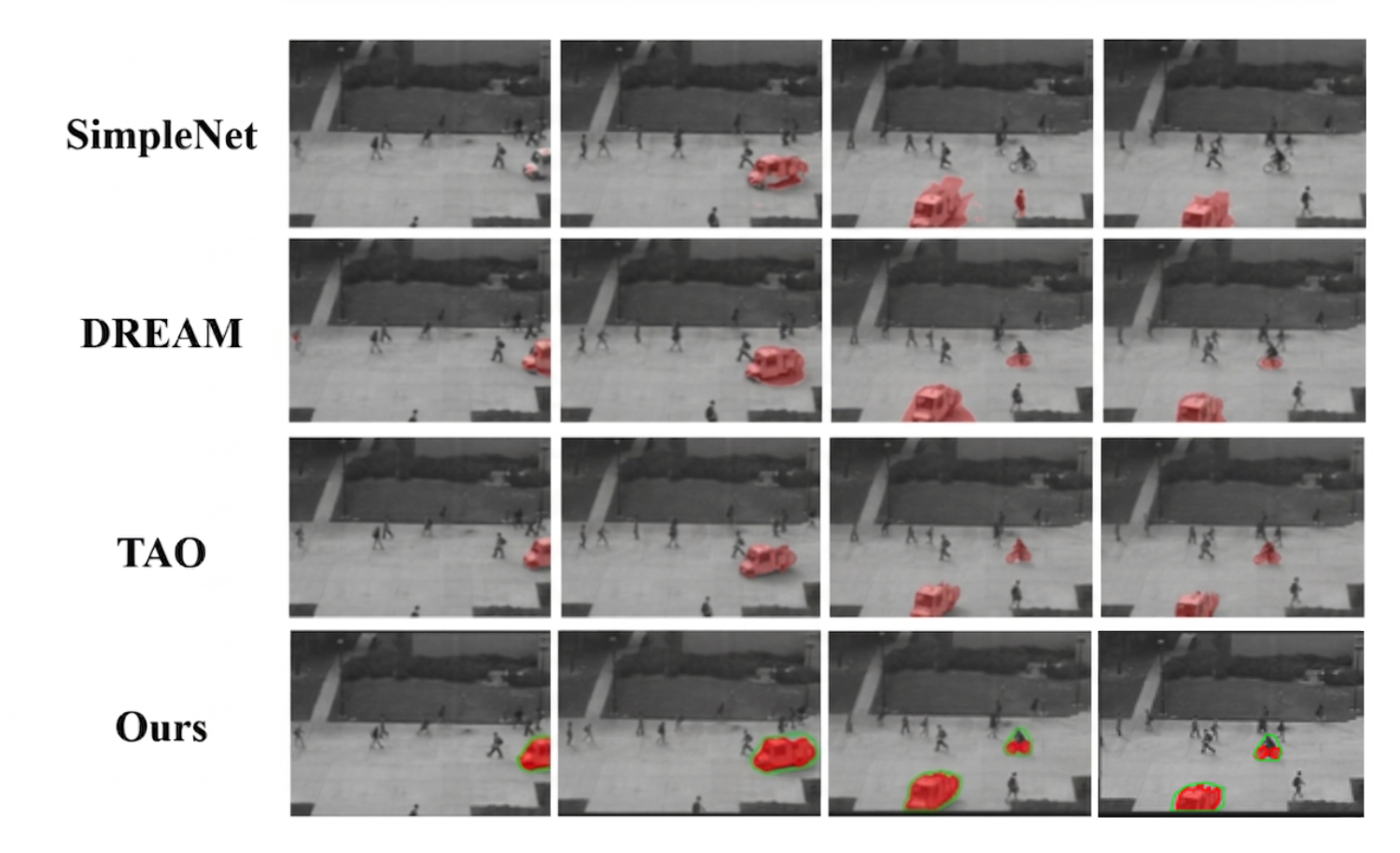} \\
\end{tabular}
}
\caption{\textbf{Qualitative comparison on the \textit{UCSD Ped2} dataset.} We compare our proposed method (\textbf{GridVAD}) against state-of-the-art baselines (\textit{SimpleNet}, \textit{DRAEM}, and \textit{TAO}) across consecutive video frames. }
\label{fig:qualitative_comparison}
\end{figure*}

%  Fig showing the "Enriched" output (Text label + SAM2 Mask).
Fig.~\ref{fig:qualitative_comparison} highlights the strengths of \textbf{\textit{GridVAD}}. The predicted masks are more complete and detailed, cleanly delineating object boundaries while preserving the integrity of anomalous objects. Notably, these qualitative gains are achieved in a zero-shot setting without dataset-specific fine-tuning.

\subsection{Discussion and Limitations}
\label{sec:limitations}

\noindent\textbf{When the proposer paradigm works.}
The results on \textit{UCSD Ped2} pixel-level metrics demonstrate that when the VLM correctly identifies an anomalous event, the downstream grounding and propagation stages produce high-quality spatiotemporal masks. The propose-ground-propagate decomposition is effective for the localization sub-problem.

\noindent\textbf{Proposal recall is the bottleneck.}
The primary failure mode is not mask quality but missed proposals. If the VLM does not generate a description for an anomaly in any of the $M$ samplings, that event produces zero score across all downstream stages. The SCC ablation (Table~\ref{tab:scc_ablation}) confirms this: disabling SCC ($M{=}1$) recovers +4.3 \textit{RBDC} and +7.3 \textit{TBDC} but degrades all pixel-level metrics. \textit{RBDC} and \textit{TBDC} measure whether GT tracks are \emph{found}, not how well they are segmented. The current pipeline deliberately prioritizes mask quality over exhaustive recall. Improving VLM proposal recall, through better prompting, adaptive re-querying of uncertain clips, or lightweight anomaly prior injection, is the most direct path to closing the gap with trained methods without sacrificing mask precision.

%\noindent\textbf{Computational cost.}
%\textit{GridVAD} utilized a VLM with 3B active parameters, which is relatively expensive compared to \textit{CLIP-based} methods. The per-clip budget is fixed at $M{+}1$ VLM calls regardless of video length, but each call is slightly heavyweight. 
\noindent\textbf{Computational cost and trade-offs.}  
By bounding VLM reasoning to a fixed $M+1$ calls per clip (avoiding frame-by-frame $\mathcal{O}(N)$ evaluation), GridVAD scales efficiently across time. However, using a 30B-parameter VLM carries a higher absolute compute footprint per clip than lightweight CLIP-based baselines. GridVAD intentionally trades per-clip inference compute for zero-shot open-set generalizability, decoupling high-level semantic proposal generation from specialized downstream execution models (\textit{Grounding DINO}, \textit{SAM 2}) that handle spatial grounding and temporal mask propagation.

\noindent\textbf{Clip independence.}
Each clip is processed independently. Anomalies that span clip boundaries may be partially missed or duplicated. Cross-clip context is currently handled only by the temporal overlap in the sliding window, not by any global reasoning.

\section{Conclusion}
\label{sec:conclusion}
 
We presented \textit{GridVAD}, a training-free pipeline that repositions VLMs from direct anomaly detectors to open-set anomaly proposers within a propose-ground-propagate decomposition. By representing video clips as stratified spatial grids, \textit{GridVAD} converts temporal anomaly detection into a single-pass image understanding task with a fixed per-clip VLM budget. Self-Consistency Consolidation filters hallucinations by retaining only proposals with cross-sampling support, while Grounding \textit{DINO} and \textit{SAM2} provide precise spatial grounding and temporal mask propagation.

Experiments on \textit{UCSD Ped2} and \textit{ShanghaiTech} Campus validate the feasibility of this decomposition: \textit{GridVAD} achieves the highest \textit{Pixel-AUROC} among all compared methods on \textit{UCSD Ped2}, including partially fine-tuned baselines, in a fully zero-shot setting. The SCC ablation reveals a controllable precision-recall tradeoff that explains the gap between strong pixel-level and lower object-level scores. Improving VLM proposal recall, through better prompting, adaptive re-querying, or lightweight anomaly priors, is the most direct path to closing this gap without sacrificing mask quality.

\bibliographystyle{splncs04}
\bibliography{main}
\end{document}